\documentclass[11pt,a4paper,twocolumn]{article}

\usepackage{titlesec}
\usepackage{color}

\usepackage[utf8]{inputenc}
\usepackage[english]{babel}
\usepackage[T1]{fontenc} % Font encoding

\usepackage{graphicx}
\graphicspath{{Images/}} % Path for images' folder
\usepackage{eso-pic} % For the background picture on the title page
\usepackage{subfig} % Numbered and caption subfigures using \subfloat
\usepackage{caption} % Coloured captions
\usepackage{float}

\usepackage{amsmath}
\usepackage{amsthm}
\usepackage{amsfonts}
\usepackage{bm}
\usepackage[overload]{empheq}  % For braced-style systems of equations

\usepackage{tabularx}
\usepackage{longtable} % tables that can span several pages
\usepackage{colortbl}
\usepackage{booktabs}

\usepackage{algorithm}
\usepackage{algorithmic}

\usepackage[colorlinks=true,linkcolor=black,anchorcolor=black,citecolor=black,filecolor=black,menucolor=black,runcolor=black,urlcolor=black]{hyperref} % Adds clickable links at references
\usepackage{cleveref}
\usepackage[square, numbers, sort&compress]{natbib} % Square brackets, citing references with numbers, citations sorted by appearance in the text and compressed
\usepackage{appendix}

\usepackage{enumitem}

\usepackage{amsthm,thmtools,xcolor} % Coloured "Theorem"
\usepackage{comment} % Comment part of code
\usepackage{fancyhdr} % Fancy headers and footers
\usepackage{lipsum} % Insert dummy text
\usepackage{tcolorbox} % Create coloured boxes (e.g. the one for the key-words)
\usepackage{stfloats} % Correct position of the tables

\newcommand{\bea}{\begin{eqnarray}} % Shortcut for equation arrays
\newcommand{\eea}{\end{eqnarray}}
\usepackage{geometry}
\definecolor{bluePoli}{cmyk}{0.4,0.1,0,0.4}

\declaretheoremstyle[
  headfont=\color{bluePoli}\normalfont\bfseries,
  bodyfont=\color{black}\normalfont\itshape,
]{colored}

\theoremstyle{colored}

\newcounter{algsubstate}

\newcolumntype{L}[1]{>{\raggedright\let\newline\\\arraybackslash\hspace{0pt}}m{#1}}
\newcolumntype{C}[1]{>{\centering\let\newline\\\arraybackslash\hspace{0pt}}m{#1}}
\newcolumntype{R}[1]{>{\raggedleft\let\newline\\\arraybackslash\hspace{0pt}}m{#1}}

\setlist[itemize,1]{label=$\bullet$}
\setlist[itemize,2]{label=$\circ$}
\setlist[itemize,3]{label=$-$}
\setlist{nosep}

\newcommand\BackgroundPic{% Adding background picture
	\put(230,358){
		\parbox[b][\paperheight]{\paperwidth}{%
			\vfill
			\centering
			\includegraphics[width=0.5\paperwidth]{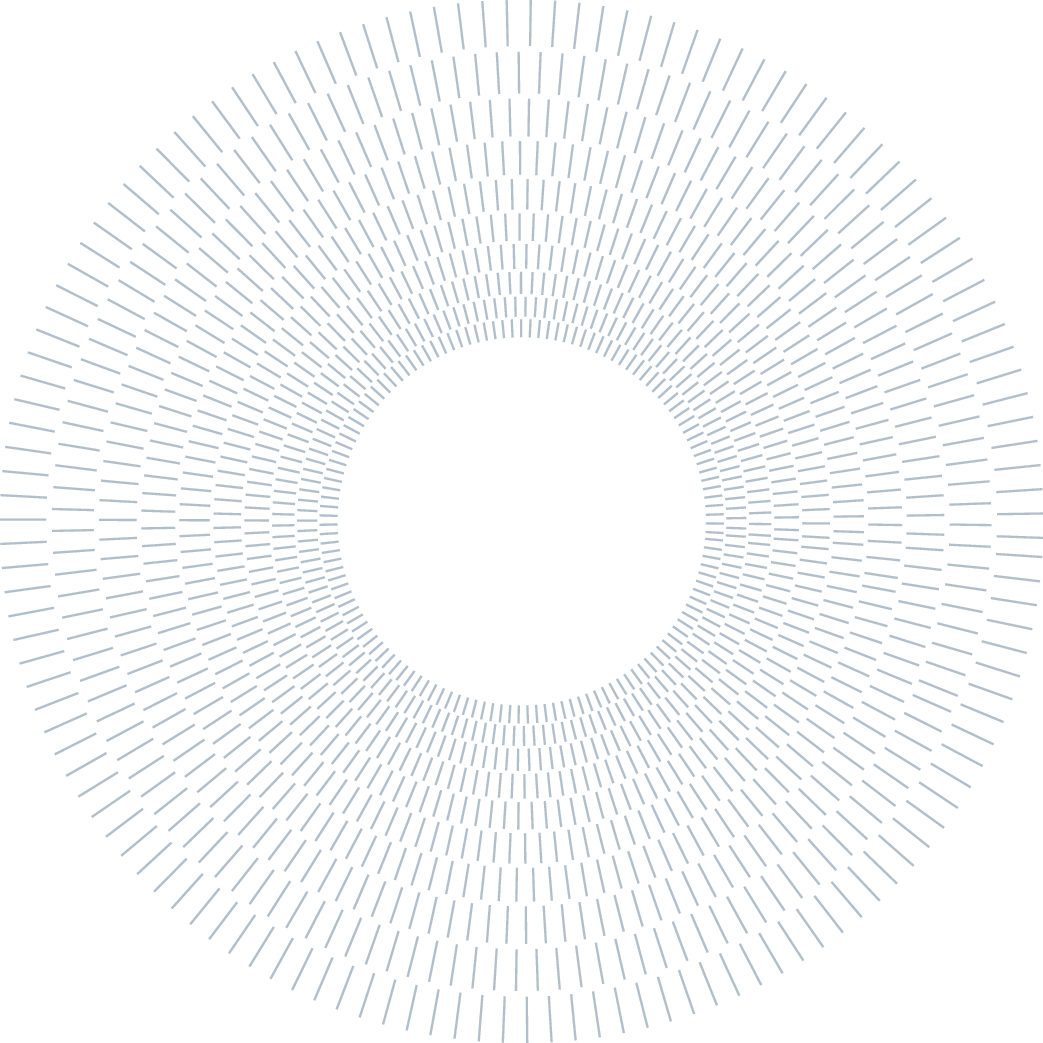}%
			\vfill
}}}

\titleformat{\section}
{\color{bluePoli}\normalfont\Large\bfseries}
{\color{bluePoli}\thesection.}{1em}{}
\titlespacing*{\section}
{0pt}{2ex}{1ex}

\titleformat{\subsection}
{\color{bluePoli}\normalfont\large\bfseries}
{\color{bluePoli}\thesubsection.}{1em}{}
\titlespacing*{\subsection}
{0pt}{2ex}{1ex}

\makeatletter
\patchcmd{\headrule}{\hrule}{\color{black}\hrule}{}{} % headrule
\patchcmd{\footrule}{\hrule}{\color{black}\hrule}{}{} % footrule
\makeatother

\renewcommand{\title}{Structure and Redundancy in Large Language Models: A Spectral Study via Random Matrix Theory}
\renewcommand{\author}{Davide Ettori}
\newcommand{\course}{Computer Science Engineering - Ingegneria Informatica}
\newcommand{\advisor}{Prof. Marco Brambilla}
\newcommand{\firstcoadvisor}{Prof. Amit Ranjan Trivedi} % insert if any otherwise comment
\newcommand{\YEAR}{2025-2026}

\begin{document}

%-----------------------------------------------------------------------------
% TITLE PAGE
%-----------------------------------------------------------------------------
% Do not change Configuration_files/TitlePage.tex (Modify it IF AND ONLY IF you need to add or delete the Co-advisors)
% This file creates the Title Page of the document
% DO NOT REMOVE SPACES BETWEEN LINES!

\twocolumn[{\begin{@twocolumnfalse}

\AddToShipoutPicture*{\BackgroundPic}

\hspace{-0.6cm}\includegraphics[width=0.6\textwidth]{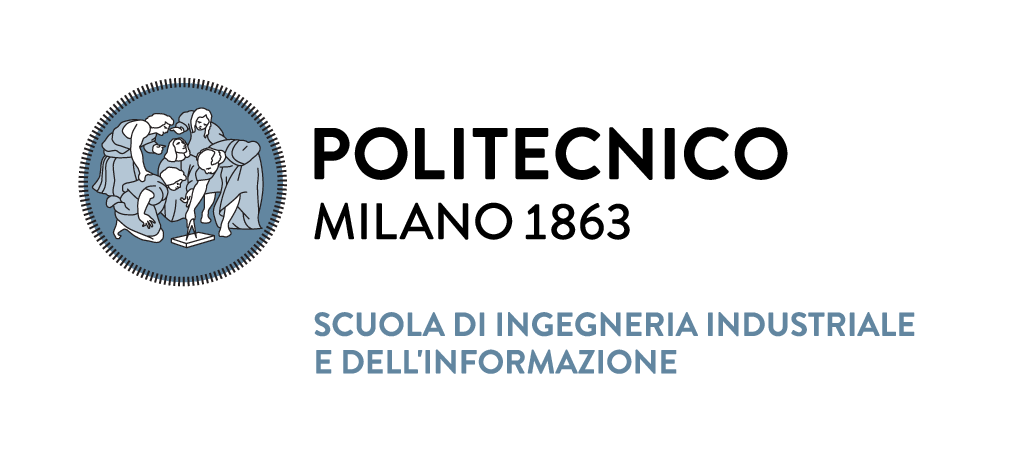}

\vspace{-1mm}
\fontsize{0.3cm}{0.5cm}\selectfont \bfseries \textsc{\color{bluePoli} Executive Summary of the Thesis}\\

\vspace{-0.2cm}
\Large{\textbf{\color{bluePoli}{\title}}}\\

\vspace{-0.2cm}
\fontsize{0.3cm}{0.5cm}\selectfont \bfseries \textsc{\color{bluePoli} Laurea Magistrale in \course}\\

\vspace{-0.2cm}
\fontsize{0.3cm}{0.5cm} \selectfont \bfseries Author: \textsc{\textbf{\author}}\\

\vspace{-0.4cm}
\fontsize{0.3cm}{0.5cm}\selectfont \bfseries Advisor: \textsc{\textbf{\advisor}}\\

% if only ONE co-advisor is present:
\vspace{-0.4cm}
\fontsize{0.3cm}{0.5cm}\selectfont \bfseries Co-advisor: \textsc{\textbf{\firstcoadvisor}}\\
% if more than one co-advisors are present:
%\vspace{-0.4cm}
%\fontsize{0.3cm}{0.5cm}\selectfont \bfseries Co-advisors: \textsc{\textbf{\firstcoadvisor}}\textsc{\textbf{\secondcoadvisor}}\\

\vspace{-0.4cm}
\fontsize{0.3cm}{0.5cm}\selectfont \bfseries Academic year: \textsc{\textbf{\YEAR}}

\small \normalfont

\vspace{11pt}

\centerline{\rule{1.0\textwidth}{0.4pt}}

\vspace{15pt}
\end{@twocolumnfalse}}]

\thispagestyle{plain} % In order to not show the header in the first page

%%%%%%%%%%%%%%%%%%%%%%%%%%%%%%
%%     THESIS MAIN TEXT     %%
%%%%%%%%%%%%%%%%%%%%%%%%%%%%%%

\section{Introduction}
\label{sec:introduction}
This thesis targets two linked challenges: reliability of large language and vision-language models (hallucinations and distribution shift) and efficiency at scale. Reliability failures erode trust, while the resource demands of large models limit deployment; a unified approach is therefore valuable. The thesis argues for principled, interpretable tools grounded in spectral geometry and Random Matrix Theory (RMT): eigenvalue spectra separate structure from noise, and the Marchenko--Pastur law plus the spiked covariance model formalize this separation into compact signatures of internal dynamics \cite{MarchenkoPastur1967,BBP2005}. Related work on hallucination detection includes black-box output checks, gray-box logit/attention uncertainty, and white-box activation probes \cite{manakul2023selfcheckgpt,mitchell2023detectgpt}. For Out-of-Distribution (OOD) detection, entropy- and energy-based scoring of logits are common baselines, alongside representation-based and spectral methods \cite{hendrycks2017baseline,Sun2022ReAct}. For efficiency, mainstream compression uses distillation, pruning, and quantization \cite{hinton2015distilling,gholami2021survey}. This thesis connect to these lines of work but emphasize spectral criteria that unify reliability and compression. A central motivation is that output-based or static uncertainty often misses failures that are visible in the evolving dynamics of internal representations. Spectral statistics provide compact, interpretable signals that can be tracked across layers and time, making them useful for both diagnosis and intervention. The first contribution, \textbf{EigenTrack} \cite{ettori2025eigentrackspectralactivationfeature}, detects hallucination and OOD behavior by tracking spectral descriptors over time with lightweight recurrent classifiers, enabling early warnings without modifying the base model. The second contribution, \textbf{RMT-KD} \cite{ettori2025rmtkdrandommatrixtheoretic}, applies the same principles to compression, projecting onto outlier eigen-directions and self-distilling the reduced model to preserve accuracy while cutting size, latency, and energy.

\section{Background}
\label{sec:background}
Random Matrix Theory (RMT) studies eigenvalue statistics of large random matrices (each entry is a random variable) and provides null models for high-dimensional noise. For symmetric i.i.d. random matrices, with variance $\sigma^2$, the Wigner semicircle law describes the limiting density, in which eigenvalues concentrate in $[-2\sigma,2\sigma]$. For covariance-type matrices, the Marchenko--Pastur (MP) law gives the asymptotic spectrum. If $X \in \mathbb{R}^{n \times p}$ has i.i.d. random entries with variance $\sigma^2$ and $p/n \to c$, then the eigenvalues of $C = \frac{1}{n}X^\top X$ lie in a bulk $[\lambda_-,\lambda_+]$ with $\lambda_\pm = \sigma^2(1 \pm \sqrt{c})^2$ and follow the distribution described by the MP law \cite{MarchenkoPastur1967}. The MP bulk is a calibrated reference for noise-like activations.
The Tracy--Widom distribution characterizes fluctuations of the largest eigenvalue at the upper spectral edge. A key framework is the spiked covariance model, where low-rank signal (spikes) is embedded in isotropic noise. The signal is: $\boldsymbol{\Sigma} = \sigma^2 \mathbf{I} + \sum_{i=1}^k \theta_i u_i u_i^T$ where the first component is noise, $\theta_i$ are spike strengths and $u_i$ are orthonormal signal directions. The Baik--Ben Arous--P{\'e}ch{\'e} (BBP) transition states that when a spike exceeds a threshold, $\theta > \sigma^2 (1+\sqrt{c})$, its sample eigenvalue detaches from the MP bulk and becomes an outlier \cite{BBP2005}. In practice, those outliers indicate structured, task-relevant directions, while the bulk is noise-like with low energy eigenvalues.

\section{Methodology: Eigentrack}
\label{sec:methodology_eigentrack}
\begin{figure}[h]
    \centering
    \includegraphics[width=0.95\linewidth]{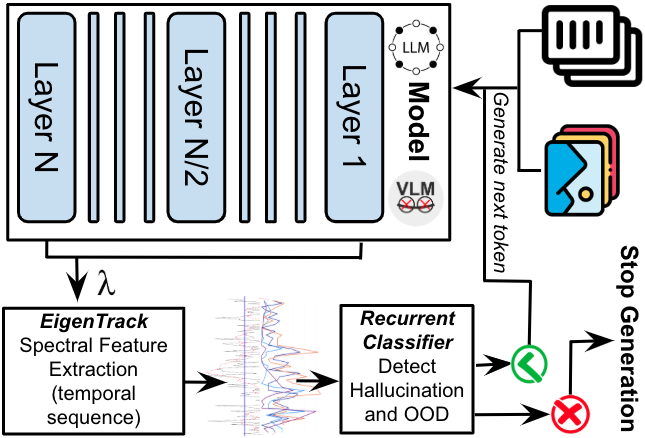}
    \caption[General architecture of EigenTrack]{General architecture of EigenTrack: spectral features extracted from hidden activations are streamed into a recurrent discrepancy detector, which outputs early warnings.}
    \label{fig:es_eigentrack_arch}
\end{figure}

EigenTrack is a real-time reliability monitor that attaches to a pretrained (Open Source) LLM or VLM and tracks how the geometry of hidden activations evolves during generation. The core hypothesis is that factual, in-distribution reasoning yields structured representations with a small number of dominant eigen-directions (consistent with the spiked covariance model), while hallucinations and OOD drift push the spectrum toward noise-like behavior (consistent with RMT), enabling the distinction. RMT provides the guiding principle: the MP bulk separates noise from informative structure, and deviations from this baseline become reliable failure indicators \cite{MarchenkoPastur1967,BBP2005}. Rather than relying on output probabilities, EigenTrack monitors the evolution of internal dynamics and therefore can flag risk early, before the model commits to a long hallucinated continuation.
At each decoding step, EigenTrack collects hidden activations from a subset of layers and concatenate them into an activation vector. The last N activation vectors are stacked into a matrix representing a sliding window of recent states. Singular Value Decomposition (SVD) yields the eigenvalues of the windowed covariance, from which a compact descriptor vector is computed, Figure~\ref{fig:es_eigentrack_arch}. The most informative descriptors include spectral entropy (dispersion), leading-eigenvalues mass (concentration), eigengaps (ratio between subsequent eigenvalues), and divergence (KL Divergence and Wasserstein distance) from the MP baseline. These descriptors are interpretable: they quantify whether the representation is collapsing toward isotropic noise or retaining structured, low-rank directions. Importantly, EigenTrack does not require access to gradients, training data, or changes to the base model; it is a monitoring head that can be attached post hoc and run concurrently with inference.

\vspace{-0.15cm}
\vspace{-0.5cm}
\begin{figure}[h]
    \centering
    \includegraphics[width=\linewidth]{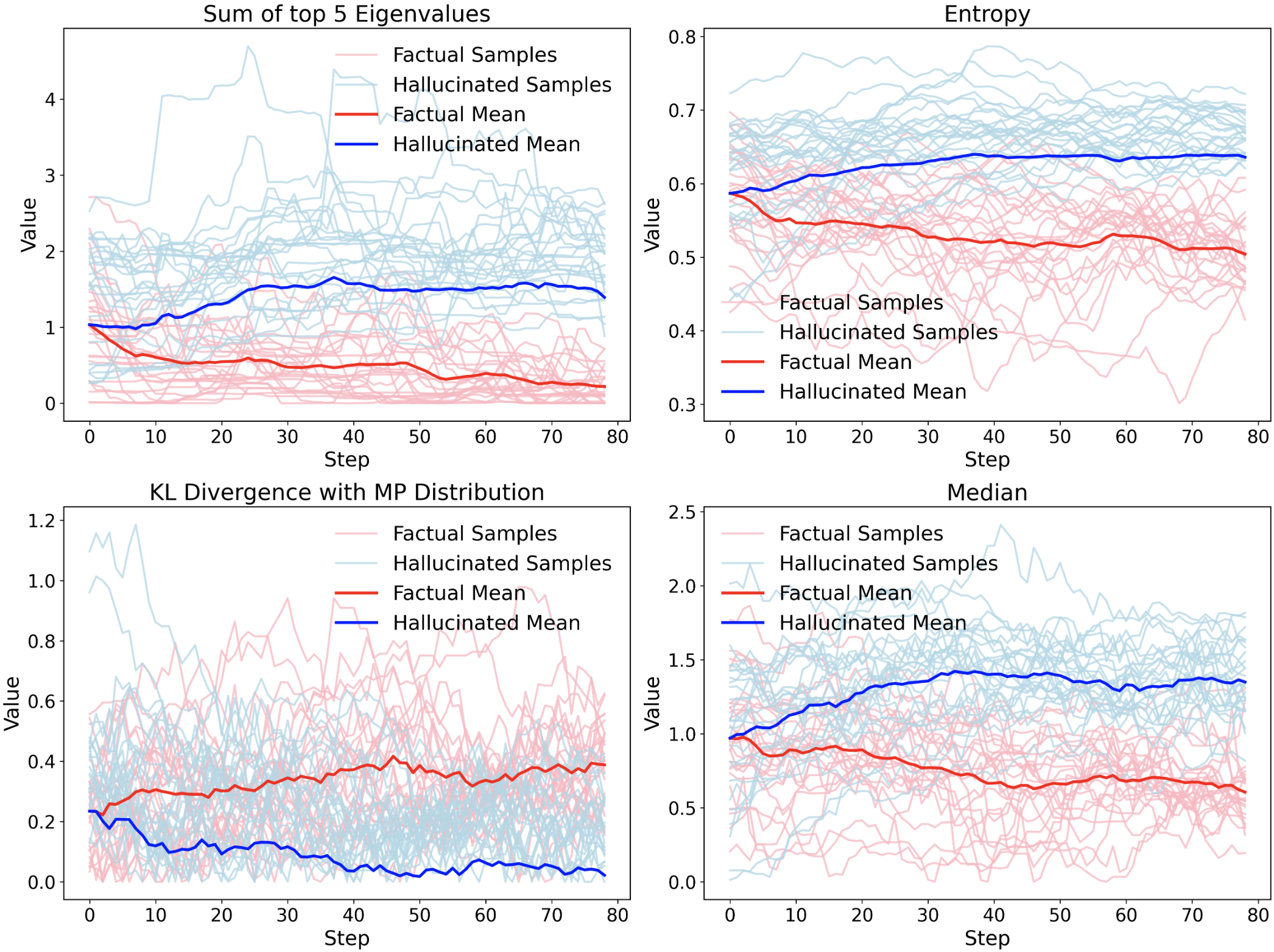}
    \caption[Temporal evolution of spectral statistics]{Temporal evolution of spectral statistics: hallucinated sequences tend to stay closer to the MP-like regime, while factual sequences diverge toward a more structured spectra.}
    \label{fig:es_eigentrack_sequences}
\end{figure}

\vspace{-0.25cm}

Single spectral snapshots can be ambiguous, so EigenTrack models the time series of descriptors with a lightweight recurrent head (RNN/GRU/LSTM). This head learns characteristic trajectories of stable versus unstable generations while remaining computationally negligible compared to the base model. The output is a per-step risk score that can trigger early intervention before hallucinated content is fully produced. The monitoring is non-invasive: the base model is unchanged and only a subset of layers is sampled, keeping overhead low and latency bounded. In practice, sampling one layer every few transformer blocks is sufficient because adjacent layers exhibit highly correlated spectral behavior. This design choice yields a favorable accuracy--latency trade-off and makes EigenTrack suitable for online deployment. From a systems perspective, EigenTrack adds minimal overhead: covariance computation uses short windows and truncated eigensolvers, the recurrent head adds only a few thousand parameters, and updates are constant per token.

\textbf{Experimental Setup:}
EigenTrack is evaluated on open-source LLMs and VLMs (LLaMa, Qwen, Mistral, LLaVa) for hallucination and OOD detection. Hallucinations are induced via a controlled QA pipeline, where context-question pairs are sampled from HotPotQA and questions are randomly swapped with unanswerable ones (where the answer does not appear in the context) 50\% of the time, and a larger LLM-as-a-judge model (e.g. LLaMa 8B) classifies whether the smaller model hallucinated. OOD evaluation contrasts in-distribution data from WebQuestions (common in web-sourced training corpora) against OOD samples from EurLex (domain-specific legal text from EU parliament).

\begin{figure}[h]
    \centering
        \includegraphics[width=\linewidth]{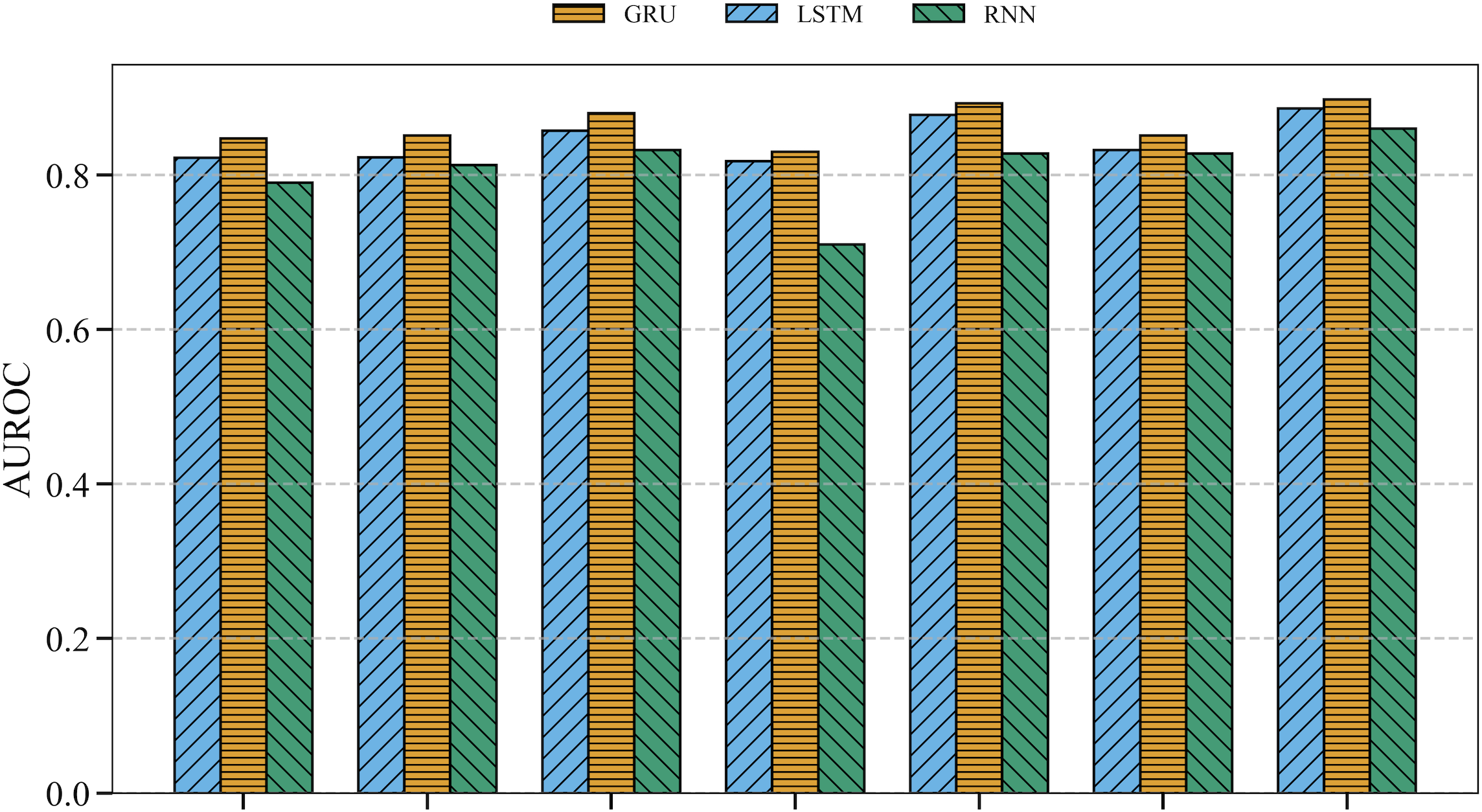}
    \\[2pt]
        \includegraphics[width=\linewidth]{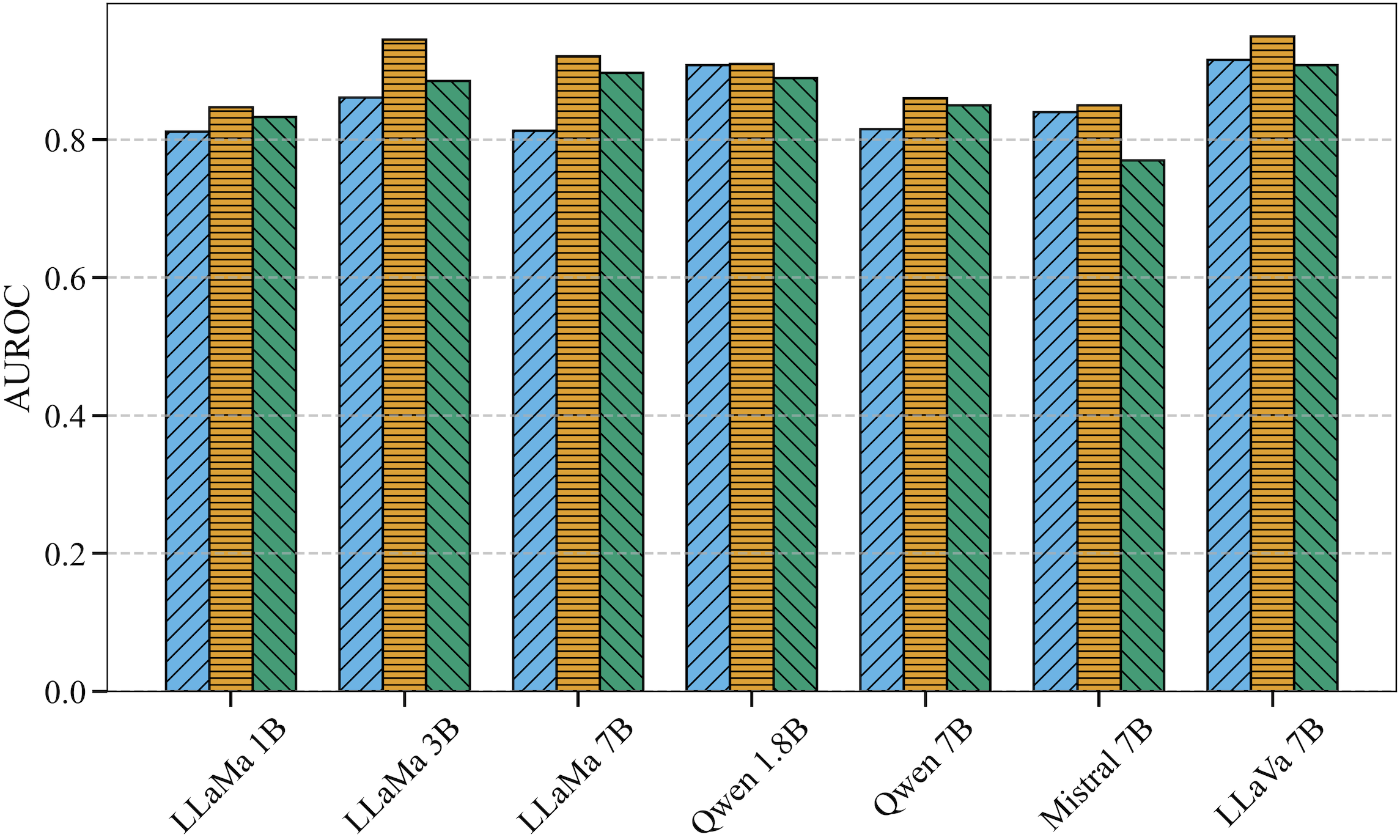}
    \caption{Hallucination \textbf{(a, top)} and OOD \textbf{(b, bottom)} detection metics across models and classifiers. Sliding window length is 30 tokens.}
    \vspace{-10pt}
    \label{fig:eigentrack_results}
\end{figure}

\textbf{Results:}
Across model families, EigenTrack consistently achieves strong AUROC for hallucination detection, with GRU heads typically outperforming RNNs and LSTMs. Performance improves with model scale, suggesting that larger models exhibit richer spectral signatures that are easier to separate. Figure~\ref{fig:eigentrack_results}a summarizes hallucination detection across architectures and recurrent heads. 
Additionally, EigenTrack achieves strong OOD performance across the same model families. Figure~\ref{fig:eigentrack_results}b shows consistent AUROC that mirror the hallucination trends, indicating that the spectral-temporal features generalize to broader distribution shifts. Beyond aggregate performance, the temporal plots in Figure~\ref{fig:es_eigentrack_sequences} show that hallucinated generations follow distinct spectral trajectories: entropy remains higher, KL divergence from the MP baseline stays lower, and eigengaps narrow, consistent with a drift toward noise-like activations. 

\begin{table}[H]
\centering
\small
\setlength{\tabcolsep}{1pt}
\renewcommand{\arraystretch}{1.3}
\caption[EigenTrack vs. baselines on LLaMa]{AUROC comparison on LLaMa for hallucination detection on HotPotQA data.}
\label{tab:es_eigentrack_sota}
\begin{tabular}{l ccc}
\toprule
\textbf{Method} & \textbf{LLaMa1B} & \textbf{LLaMa3B} & \textbf{LLaMa7B} \\
\midrule
\textbf{EigenTrack}  & \textbf{0.842} & \textbf{0.861} & \textbf{0.894} \\
LapEigvals           & 0.785 & 0.819 & \underline{0.871} \\
INSIDE               & 0.753 & \underline{0.831} & 0.810 \\
SelfCheckGPT         & 0.739 & 0.804 & 0.809 \\
HaloScope            & \underline{0.820} & 0.827 & 0.861 \\
\bottomrule
\end{tabular}
\end{table}

\vspace{-0.2cm}

These dynamics explain why temporal modeling is crucial: risk emerges as a gradual spectral drift rather than a single anomalous point. Overall, EigenTrack indicates that spectral dynamics provide early, interpretable failure signals while remaining lightweight, supporting the thesis claim that RMT-based geometry can ground practical reliability tools. Table~\ref{tab:es_eigentrack_sota} summarizes state-of-the-art hallucination detection on the LLaMa family, with additional comparisons mentioned in the thesis document.

\begin{figure}[h]
    \centering
        \includegraphics[width=\linewidth]{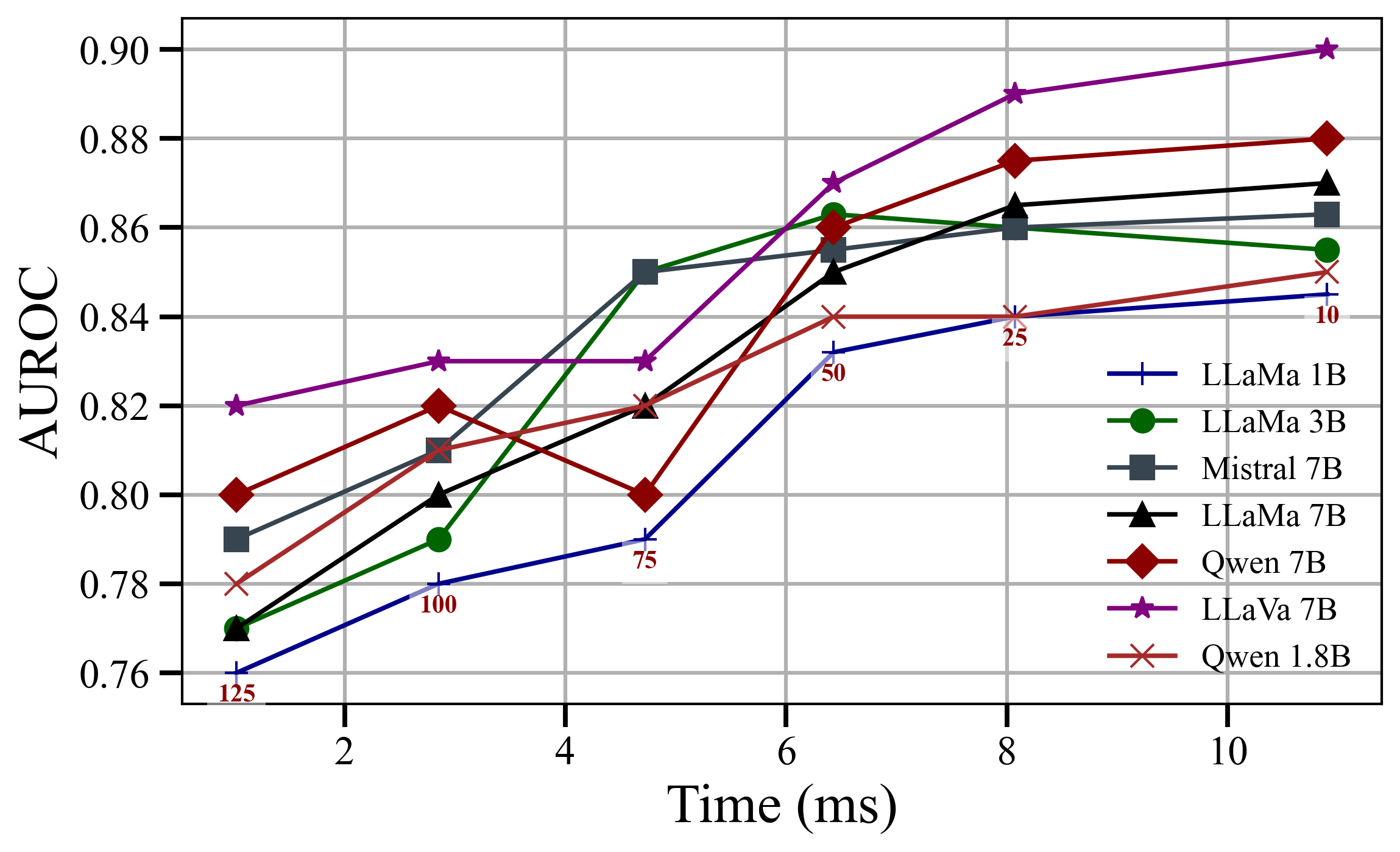}
    \\[2pt]
        \includegraphics[width=\linewidth]{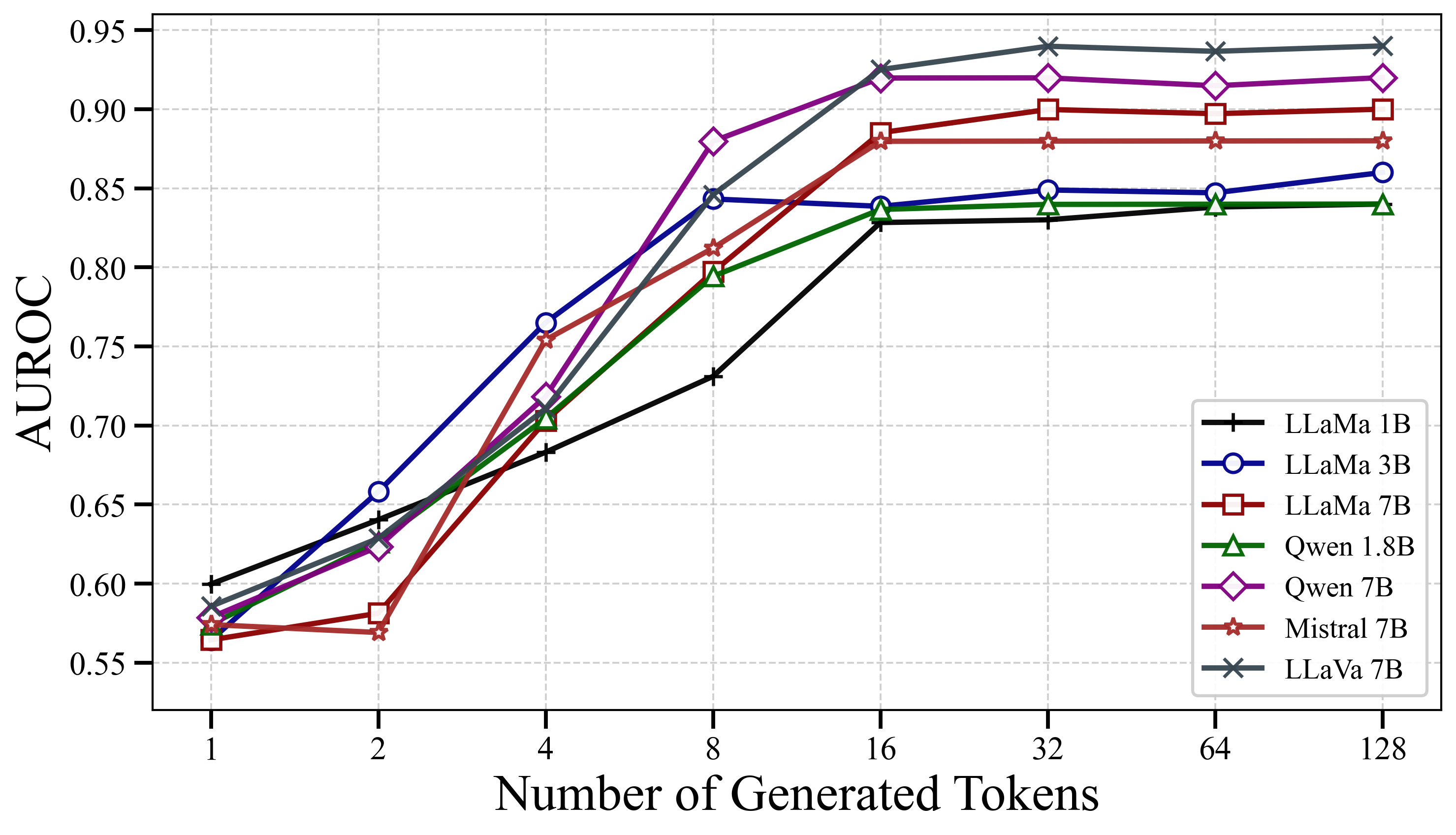}
    \caption{AUROC and latency as a function of sequence window length of GRU on hallucination dataset \textbf{(a, top)} and AUROC as a function of total generated tokens \textbf{(b, bottom)}.}
    \vspace{-10pt}
    \label{fig:eigentrack_ablation}
\end{figure}

\textbf{Ablation Study:}
Figure~\ref{fig:eigentrack_ablation}a shows AUROC and inference latency versus sliding-window length. Short windows yield higher AUROC by capturing fine-grained dynamics but increase latency due to more evaluations; longer windows reduce latency while gradually degrading performance. A clear knee at 25 tokens offers the best trade-off. Below this range, spectral estimates are noisy and latency rise; above it, excess context oversmooths signals and degrades detection, indicating that hallucination dynamics lie within a moderate temporal horizon. Figure~\ref{fig:eigentrack_ablation}b plots AUROC against the number of observed tokens. Performance rises sharply from chance within the first few tokens and quickly saturates, showing that hallucination cues emerge early, with diminishing gains from later tokens, supporting early stopping to balance accuracy and latency. This indicates that hallucination isn't a sudden event and it can be detected, from internal activations, before manifesting in the text output.

\vspace{-0.2cm}

\section{Methodology: RMT-KD}
\label{sec:methodology_rmtkd}
\begin{figure}[h]
    \centering
    \includegraphics[width=\linewidth]{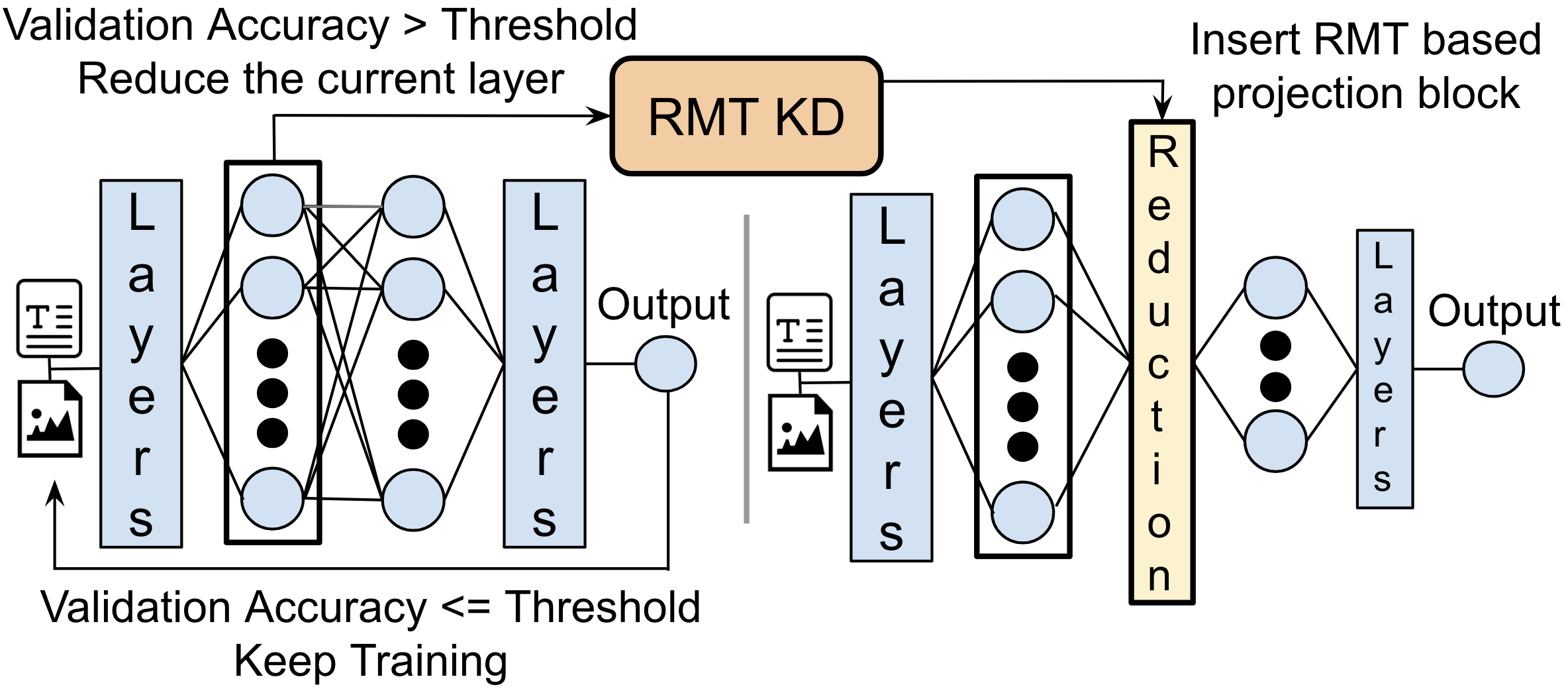}
    \caption{Overview of the iterative RMT-KD pipeline: spectral analysis estimates the MP bulk edge, outlier eigenvectors define the causal subspace, and self-distillation stabilizes training.}
    \label{fig:es_rmtkd_arch}
\end{figure}

RMT-KD is a compression framework that uses RMT to identify and preserve only the causal directions of hidden activations, yielding smaller models without sacrificing accuracy.
The idea is that activation spectra exhibit a noise bulk predicted by the MP law, while task-relevant structure appears as outlier eigenvalues beyond the bulk edge; those outliers define the subspace worth keeping \cite{MarchenkoPastur1967,BBP2005}. 
RMT-KD, shown in Figure~\ref{fig:es_rmtkd_arch}, operationalizes this into a staged procedure: train until a target level of accuracy is achieved (so we know there is signal in the model), analyze first layer activations on a calibration subset, estimate the upper MP bulk edge, project onto the outlier eigenvectors subspace, self-distill to recover accuracy and repeat for each layer until target compression is met.

\vspace{-0.2cm}

\begin{figure}[h]
    \centering
    \includegraphics[width=\linewidth]{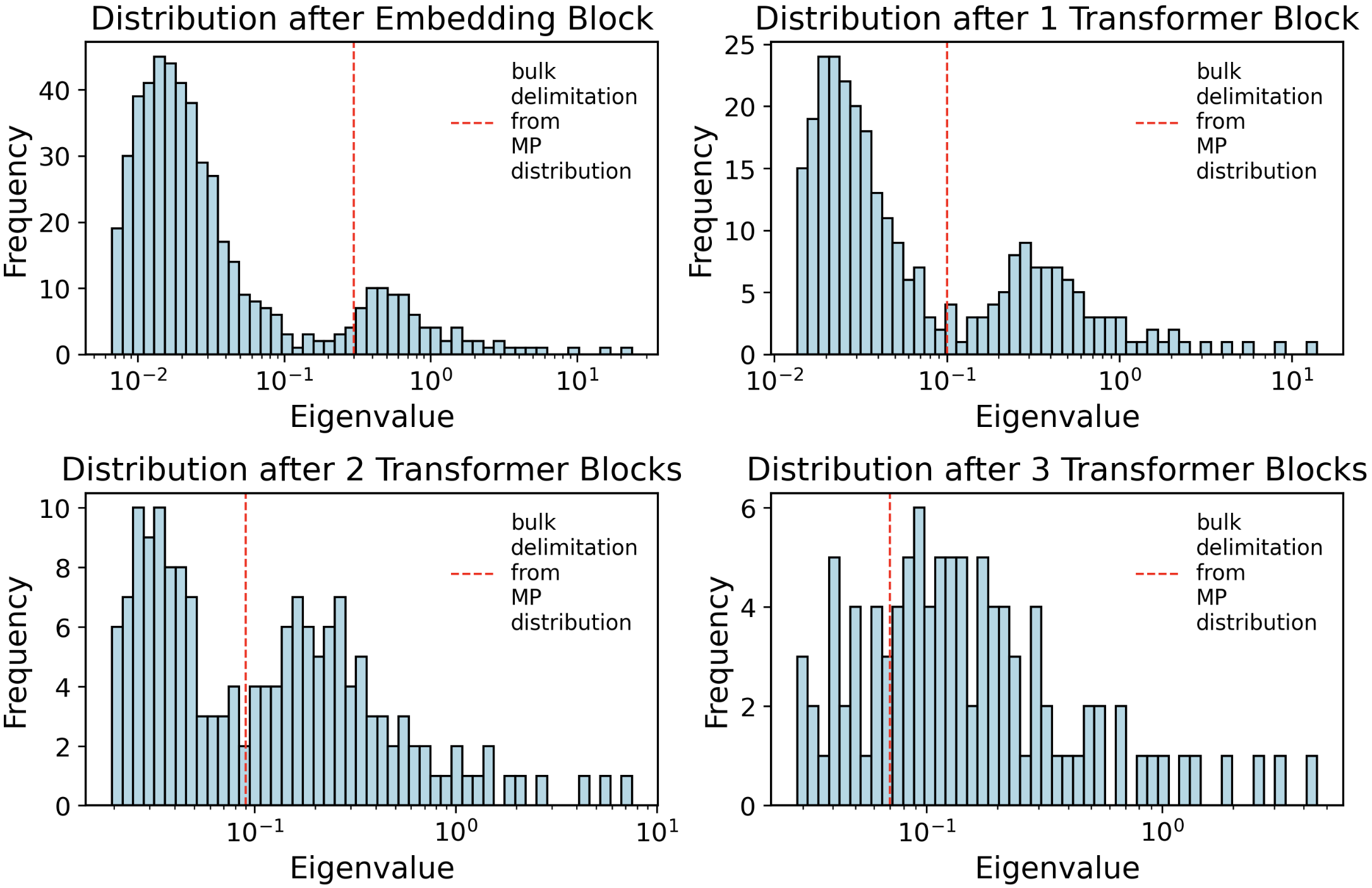}
    \caption{Evolution of the empirical eigenvalue distribution of activation covariances across layers of BERT on SST data during training.}
    \label{fig:eig_dist}
\end{figure}

At each reduction step, activations from a target layer are collected to form a covariance matrix. The eigenvalue spectrum is matched to the MP density to estimate the noise variance and its upper edge \(\lambda_+\). Eigenvalues above \(\lambda_+\) are treated as signal, and their eigenvectors define a projection that reduces the layer width while preserving the most informative directions, as shown in Figure~\ref{fig:eig_dist}. Because the projection is dense, the resulting model remains hardware-friendly and does not require sparse kernels. 

\vspace{-0.2cm}
\vspace{-10pt}

Self-distillation is the stabilizing step: after each projection, the reduced model (student) is fine-tuned to match the logits of the pre-reduction checkpoint (teacher) while still optimizing the task loss. This “teacher-from-previous-stage” scheme transfers decision boundaries into the smaller subspace, preventing catastrophic forgetting and allowing aggressive reductions to accumulate layer by layer. The process is modular and can be applied to transformers and CNNs with minimal architectural modifications.

\textbf{Experimental Setup:}
RMT-KD is evaluated on BERT-base and BERT-tiny across GLUE tasks (SST, QQP, QNLI) and on ResNet-50 for CIFAR-10. Models share training settings; only the RMT projections and self-distillation schedule vary. A small calibration subset is used for spectral estimation, and performance is tracked in accuracy, parameter reduction, throughput, power, memory, and energy per inference.

\begin{figure}[h]
    \centering
        \includegraphics[width=\linewidth]{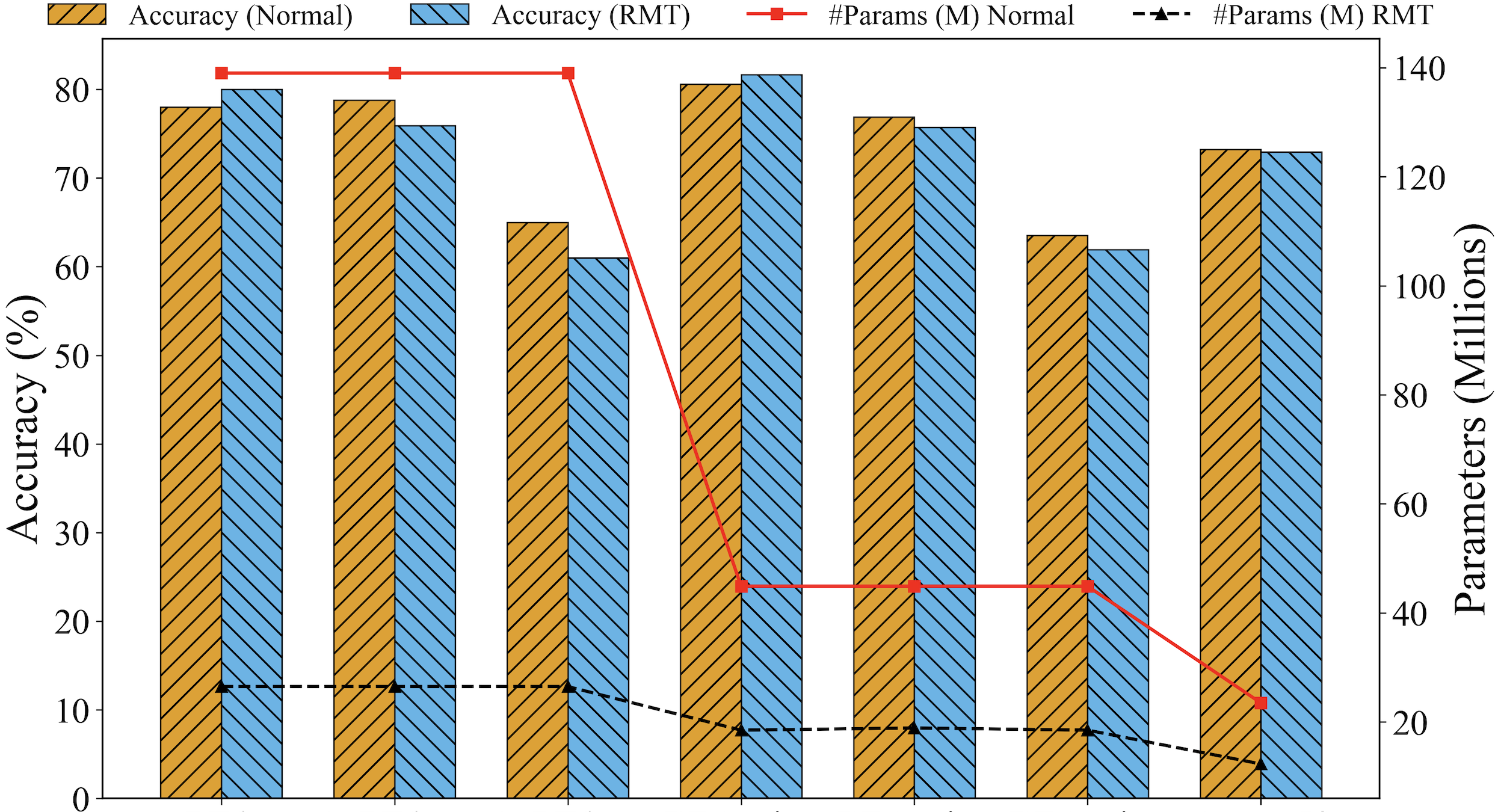}
    \\[2pt]
        \includegraphics[width=\linewidth]{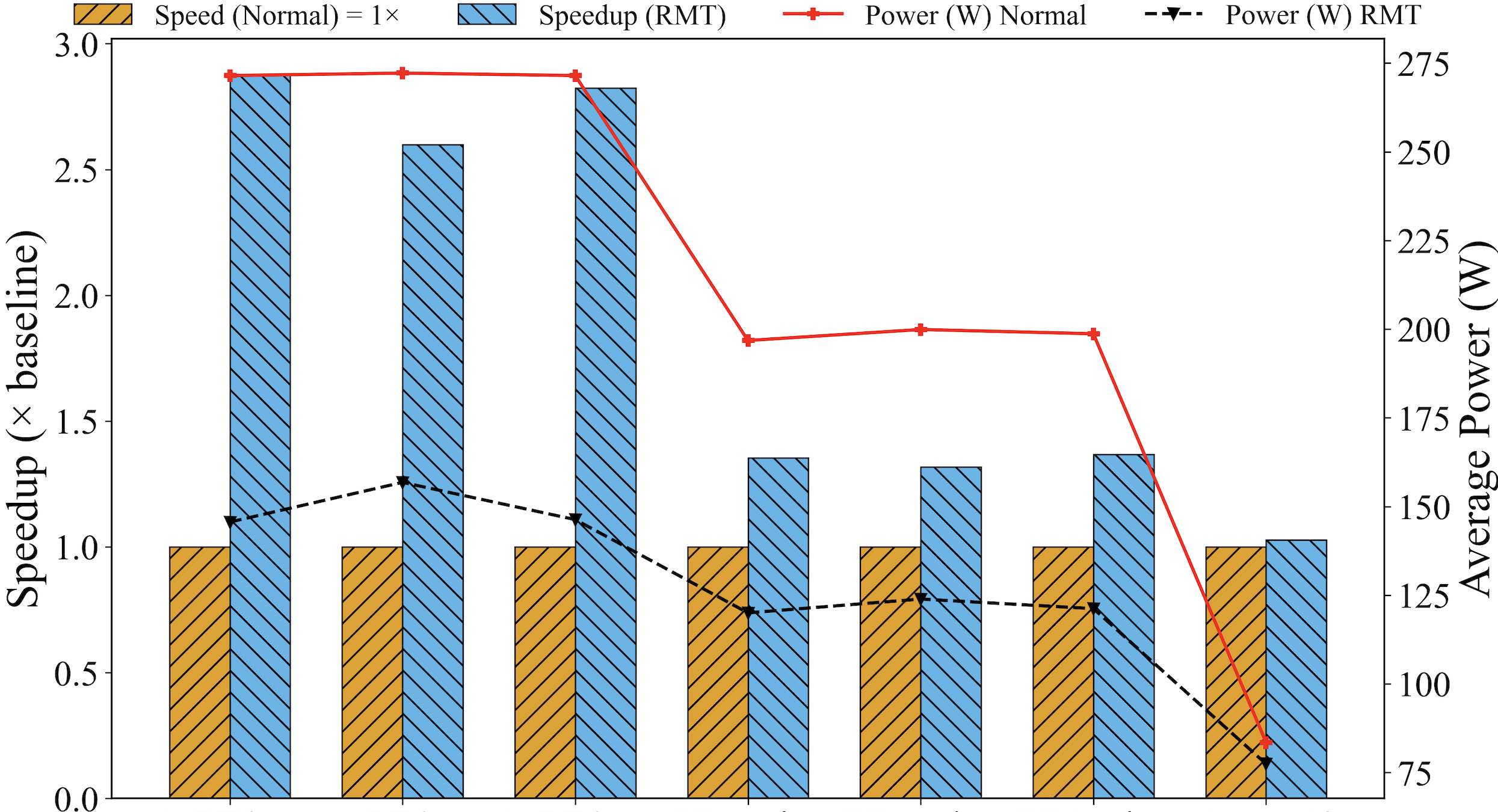}
    \\[2pt]
        \includegraphics[width=\linewidth]{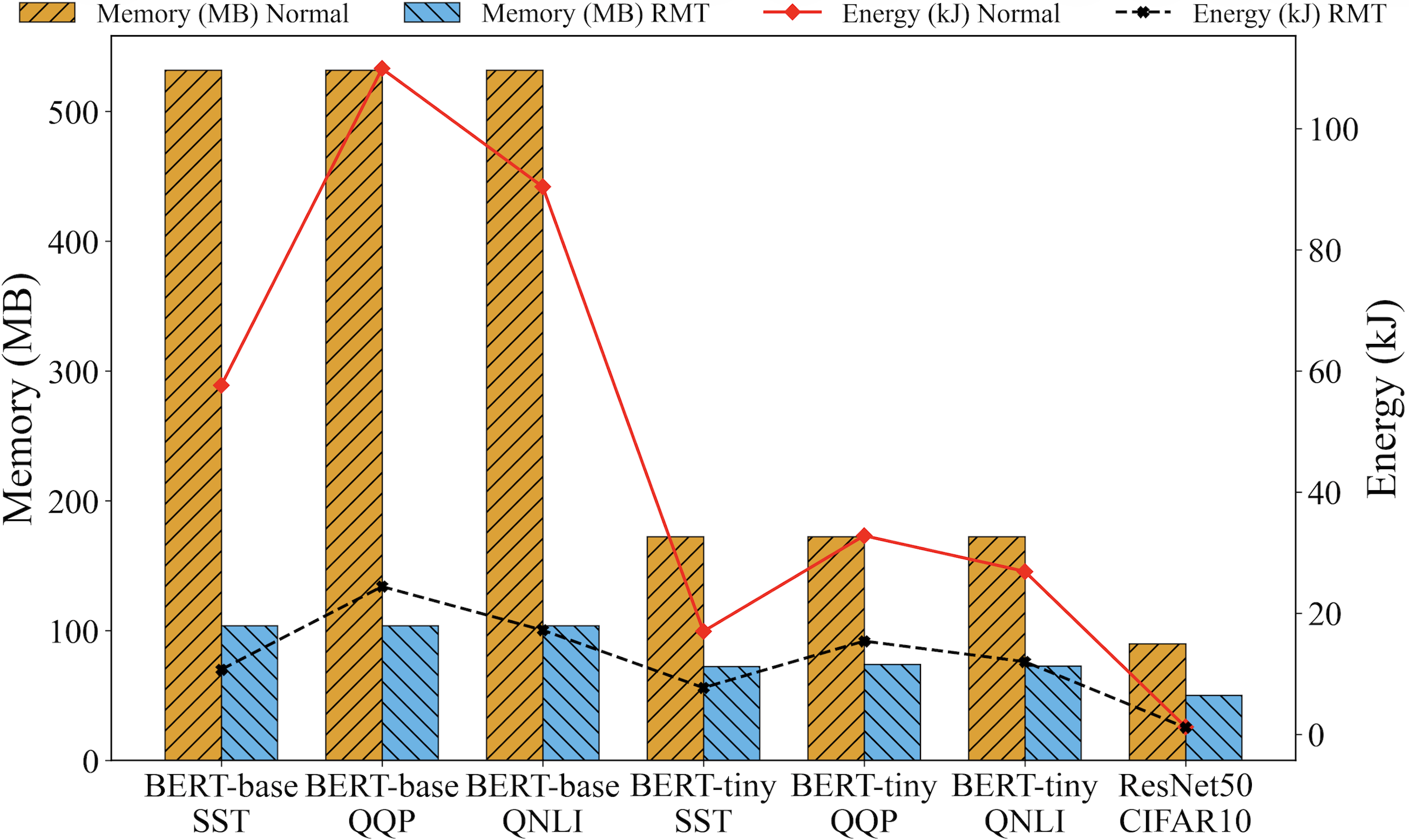}
    \caption{Performance metrics of RMT-KD models relative to baselines: \textbf{(a, top)} accuracy and parameter reduction, \textbf{(b, middle)} inference speedup and power reduction, \textbf{(c, bottom)} memory footprint and energy per inference.}
    \vspace{-10pt}
    \label{fig:rmtkd_results}
\end{figure}

\textbf{Results:}
RMT-KD achieves large parameter reductions while maintaining or improving accuracy. Figure~\ref{fig:rmtkd_results}a shows the accuracy–compression trade-off: BERT-base sees reductions around 80\% with slight accuracy gains on SST/QQP, BERT-tiny retains accuracy with ~60\% reduction, and ResNet-50 achieves nearly 50\% reduction with minimal loss. The pattern suggests that spectral filtering removes redundant or noisy directions and can act as a regularizer, improving generalization even as capacity shrinks. The gains are consistent across architectures, indicating that the spectral criterion captures a general property of learned representations rather than a task-specific artifact.
Efficiency gains are reflected in Figure~\ref{fig:rmtkd_results}b: throughput increases (nearly 3× on BERT-base in some tasks) while power consumption drops, indicating real hardware-level benefits from dimensionality reduction. Figure~\ref{fig:rmtkd_results}c complements this view by showing that memory footprint and total energy per inference decrease substantially, especially for larger models. 
These results matter for deployment because they reduce both latency and operational cost, and they do so without introducing sparsity; the compressed models remain dense and therefore align well with standard GPU kernels. This confirms that spectral projection is a principled alternative to heuristic pruning or low-rank factorization, preserving dense computation while cutting cost.

\begin{table}[h]
\centering
\small
\setlength{\tabcolsep}{6pt}
\renewcommand{\arraystretch}{1.2}
\caption[RMT-KD vs. baselines]{Compression ratio (Red.) and accuracy change (Acc.) comparison with other methods.}
\label{tab:es_rmtkd_sota}
\resizebox{\linewidth}{!}{
\begin{tabular}{lcccccc}
\toprule
& \multicolumn{2}{c}{\textbf{BERT-base}} & \multicolumn{2}{c}{\textbf{BERT-tiny}} & \multicolumn{2}{c}{\textbf{ResNet-50}} \\
\cmidrule(lr){2-3}\cmidrule(lr){4-5}\cmidrule(lr){6-7}
\textbf{Method} & \textbf{Red.} & \textbf{Acc.} & \textbf{Red.} & \textbf{Acc.} & \textbf{Red.} & \textbf{Acc.} \\
\midrule
\textbf{RMT-KD} & \textbf{80.9\%} & \textbf{+1.8\%} & \textbf{58.8\%} & \textbf{+1.4\%} & \textbf{47.7\%} & \textbf{+0.7\%} \\
DistilBERT & 42.7\% & +0.2\% & \underline{54.8\%} & \underline{+0.4\%} & \multicolumn{2}{c}{\textemdash} \\
Theseus & \underline{48.3\%} & \underline{+0.6\%} & 53.0\% & +0.1\% & \multicolumn{2}{c}{\textemdash} \\
PKD & 40.5\% & -1.0\% & 50.1\% & -0.8\% & \multicolumn{2}{c}{\textemdash} \\
AT & \multicolumn{2}{c}{\textemdash} & \multicolumn{2}{c}{\textemdash} & 42.2\% & +0.4\% \\
FitNet & \multicolumn{2}{c}{\textemdash} & \multicolumn{2}{c}{\textemdash} & 40.6\% & +0.2\% \\
CRD & \multicolumn{2}{c}{\textemdash} & \multicolumn{2}{c}{\textemdash} & \underline{45.4\%} & \underline{+0.6\%} \\
\bottomrule
\end{tabular}
}
\end{table}

Table~\ref{tab:es_rmtkd_sota} provides a compact state-of-the-art comparison against representative distillation baselines (data for BERT are the average performance between SST, QQP and QNLI,  while CIFAR10 is used for ResNet); additional metrics and comparisons are reported in the thesis. The takeaway is that RMT-KD achieves both stronger compression and competitive performance, remaining dense and hardware-friendly.

\begin{figure}[h]
    \centering
    \includegraphics[width=\linewidth]{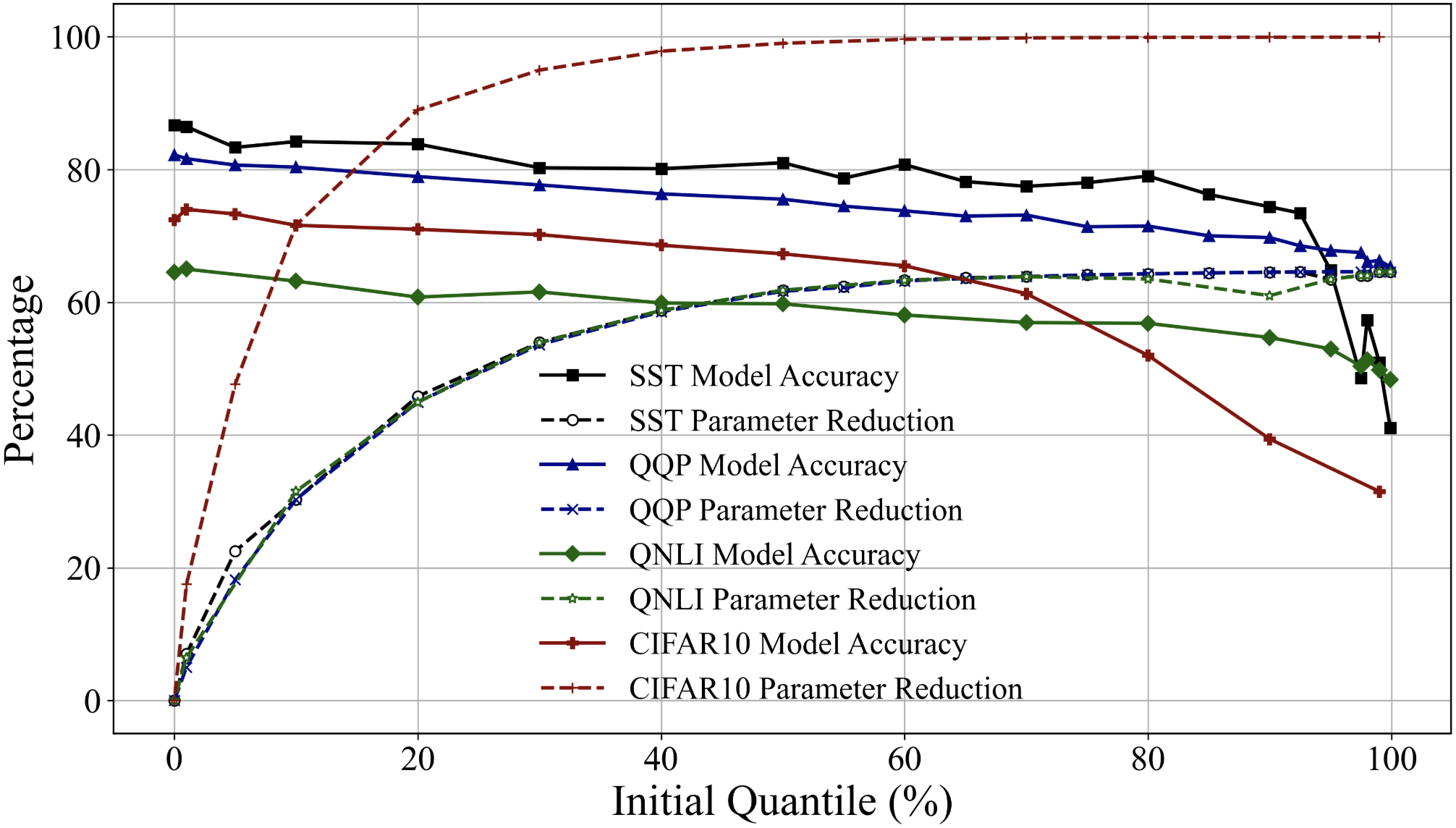}
    \caption[Quantile ablation]{Effect of the variance initialization quantile on the compression-accuracy tradeoff.}
    \label{fig:es_rmtkd_ablation}
\end{figure}

\textbf{Ablation Study:}
Compression aggressiveness is controlled by the quantile used to initialize the MP law variance. That is the expected variance in the activations matrix data, estimated as a quantile of the empirical eigenvalue distribution of the covariance matrix. Figure~\ref{fig:es_rmtkd_ablation} shows that moderate quantiles (around the median) yield the best accuracy–compression trade-off, while higher quantiles become overly aggressive and degrade accuracy. This provides a tunable parameter balancing efficiency and performance. ResNet offers greater reduction potential, while BERT is constrained by fixed embedding and token projection layers, which dominate beyond 40\% quantile and limit further compression.

\vspace{-0.7cm}

\section{Conclusions}
This thesis shows that spectral geometry and RMT offer a unified view for improving both reliability and efficiency in deep learning. EigenTrack provides early, interpretable warnings of hallucination and OOD behavior without changing the base model. RMT-KD preserves causal eigen-directions via dense projection and self-distillation, achieving strong compression with minimal accuracy loss and clear system-level gains. These contributions position eigenvalue dynamics as a shared language for diagnosis and optimization.
Limitations include evaluation scope and the cost of spectral computation for very large layers. Future work should scale to larger multimodal models, explore hybrid systems combining RMT analysis on attention matrices, and integrate approximate eigensolvers.

%\section{Acknowledgements}
%\input{sections_summary/ack.tex}

%\clearpage

\vspace{-0.2cm}
%\bibliography{bibliography.bib}

\end{document}